\documentclass[twocolumn]{svjour3}

\usepackage{cs}
\usepackage{mathsymb}
\usepackage{algorithm}
\usepackage{algorithmic}
\usepackage{graphicx}
\usepackage{url}
\usepackage{cases}
\usepackage{subfigure}
\usepackage{diagbox}
\graphicspath{{./}{./Fig/}{./Figs/}}

\renewcommand{\algorithmicrequire}{\textbf{Input:}}
\renewcommand{\algorithmicensure}{\textbf{Output:}}

\usepackage{rotating}

\begin{document}

\title{Reading Car License Plates
      Using Deep Convolutional Neural Networks and LSTMs}

\author{Hui Li,
        Chunhua Shen
\thanks{
  The authors are with School of Computer Science,
  The University of Adelaide, Australia; and  Australian Centre for Robotic Vision.
  Correspondence should be addressed to C. Shen (email: chunhua.shen@adelaide.edu.au).
}
}

\institute{}

\markboth{Manuscript}%
{Li \MakeLowercase{\textit{et al.}}:
      Reading Car License Plates
      Using Deep Convolutional Neural Networks and LSTMs
}

\maketitle

\begin{abstract}

In this work, we tackle the problem of car license plate detection and recognition in
natural scene images.
Inspired by the success of deep  neural networks (DNNs)
in various vision applications,
here we leverage
DNNs to learn high-level features in a cascade framework, which lead
to improved performance on both detection and recognition.

Firstly, we train a $37$-class convolutional neural network (CNN) to detect all characters in an image, which
results in a high recall, compared with conventional approaches such as training a binary
text/non-text classifier.  False positives
are then eliminated by the second  plate/non-plate
CNN classifier. Bounding box refinement is then carried out based on the edge
information of the license plates, in order to improve the intersection-over-union (IoU)
ratio. The proposed cascade framework extracts license plates effectively with both
high recall and precision.
Last, we propose to recognize the license characters as a
{\em sequence labelling} problem.
A recurrent neural network (RNN) with long short-term
memory (LSTM) is trained to recognize the sequential features extracted from
the whole license plate via CNNs.
The main advantage of this approach is that it is segmentation free.
By exploring context information and avoiding
errors caused by segmentation, the RNN method performs better than
a baseline method of combining segmentation and deep CNN classification; and
achieves state-of-the-art recognition accuracy.


\end{abstract}

\section{Introduction}
\label{sec:intro}

\begin{figure*}[ht!]
\centering
\includegraphics[width=0.8\textwidth]{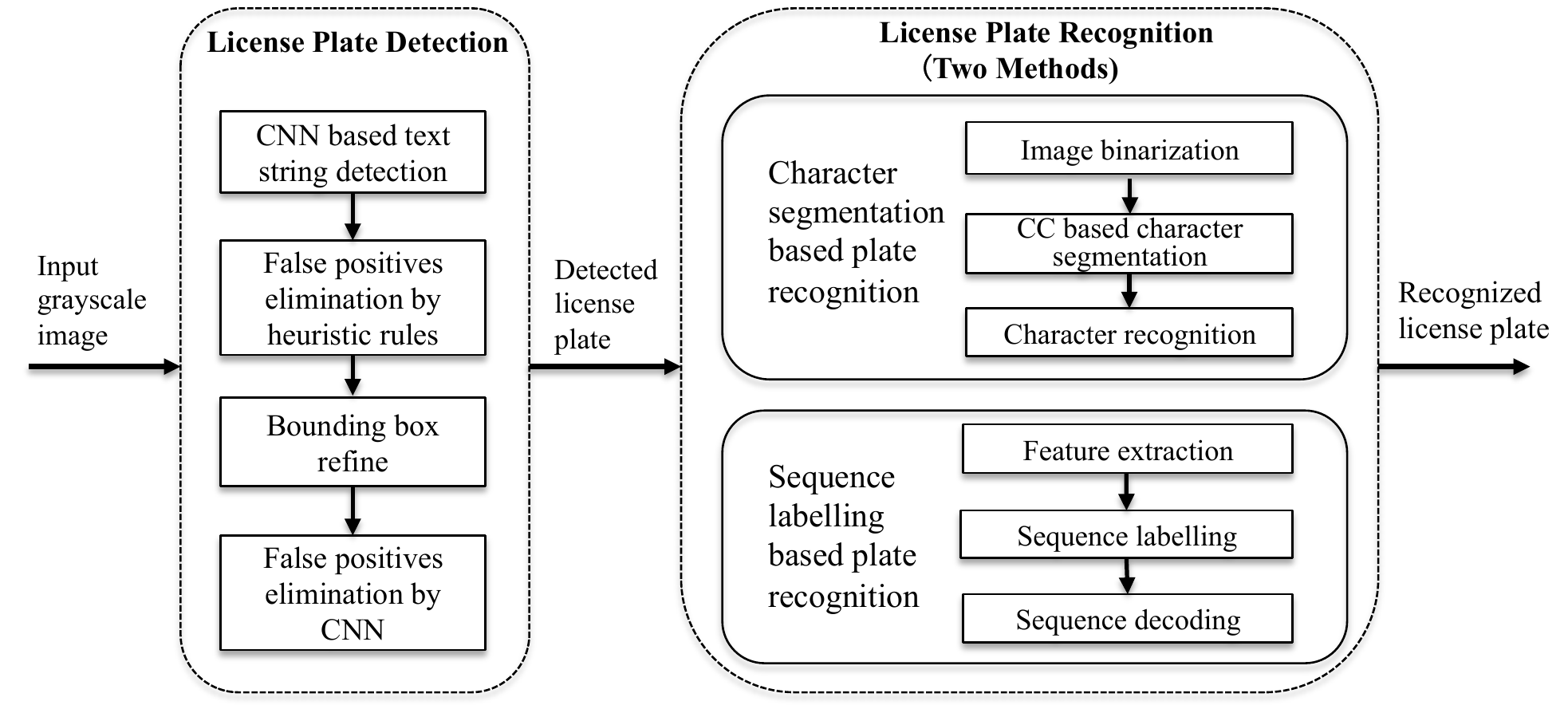}
\caption{The overall framework of the proposed
  car license plate detection and recognition method.
  Here the license plate recognition part shows two independent methods:
  the first one is the baseline (segmentation based) and the bottom one is the LSTM based approach.
}
\label{Fig:1}
\end{figure*}

With the recent advances in intelligent transportation systems, automatic car license plate detection and recognition (LPDR)
has attracted considerable research interests.
It has a variety of potential applications, such as in security and traffic control.
Much work has been done on the topic of LPDR.

However, most of the existing algorithms work well only under controlled conditions.
For instance, some systems require sophisticated hardware to capture high-quality images,
and others demand vehicles to slowly pass a fixed access gate   or even at a full stop.
It is still a challenging task
to detect license plate and recognize its characters accurately in an open environment.
The difficulty lies in the extreme diversity of character patterns, such as different sizes,
fonts and colors across nations, character distortion caused by capturing viewpoint,
and low-quality images caused by uneven lighting, occlusion or blurring.
The highly complicated background makes the problem even intricate,
especially the general text in shop boards, text-like outliers like windows, guardrail, bricks, which often lead to false alarms in detection.

A complete LPDR system is typically divided into two subsequent components:
detection and recognition. Plate detection means to localize the license plate and generate suitable
bounding boxes, while plate recognition aims to identify the characters depicted within the  b\-oun\-d\-ing
boxes.

Previous work on license plate detection usually relies
on some hand\-crafted image features that capture certain morphological,
color or textural attributes of the license plate~\cite{Du2013Automatic,Cnagnost2008}.
These features can be sensitive to image noises, and may result in many
false positives under complex backgrounds or under different illumination conditions.
In this paper, we tackle those problems by leveraging the high capability of convolutional neural networks (CNNs).
CNNs have demonstrated impressive performance on various tasks including image classification, object detection,
semantic segmentation, \etc~\cite{Ross2014}. A CNN consists of multiple layers of  neurons,
which can learn high-level features efficiently from a large amount of labeled training data.
In our LPDR system, we view license plates as strings of text.
A cascade framework is proposed which examines the image for the presence of text firstly,
and then reject text that is not license plates. Both stages use
CNN classifiers that show high discriminative ability and strong robustness against complicated backgrounds.
As for license plate recognition, we present two methods.
The first one serves as a baseline, which
segments the characters from the detected license plate beforehand,
and recognizes each character using CNNs.

More importantly, for the second method, {\em
we propose to recognize the license characters as a sequence labelling problem}.
The car  plate image is viewed as an unsegmented sequence.
CNNs are used to extract  image features. A recurrent neural
network (RNN) with connectionist temporal classification
(CTC)~\cite{Graves2009Pami} as the output layer is employed to label the
sequential data. With this method, we do not need to deal with the challenging
character segmentation task. The recurrent property of RNNs also
helps to exploit the context information and improve the recognition
performance. The overall framework of our LPDR system is shown in
Fig.~\ref{Fig:1}, including both recognition approaches.
Thus the main contributions of this work are as follows.

\begin{itemize}
  \item

  We propose a cascade framework that uses different CNN classifiers for
  different purpose. To begin with, a $4$-layer $37$-class ($10$ digits, $26$ uppercase letters plus the negative non-character category) CNN classifier is employed in a sliding-window fashion across the entire image to detect
  the presence of text and generate a text saliency map.  Text-like regions are
  extracted based on the clustering nature of characters. Then another
  $4$-layer plate/non-plate CNN classifier is adopted to reject false positives
  and distinguish license plates from general text. With this framework, our
  system can detect license plates in complicated backgrounds with high recall
  and precision. Moreover, it can be used to detect license plates of different styles (e.g., from different
  countries), regardless of different plate colors, fonts or sizes.

\item

  We train another $9$-layer CNN classifier for the second-stage character recognition.
  The deeper CNN architecture possesses much stronger representation ability,
  and can improve the recognition performance significantly. In addition, we
add local binary pattern (LBP) features into the CNN input, on top of the original gray-scale channel,
and fine-tune the $9$-layer CNN model.  Experimental results show enhanced classification performance with the additional LBP feature channels.
The final character recognition combines the prediction results of the two CNN classifiers, which leads to improved results.

\item In order to overcome the difficulty in character segmentation, we develop a deep recurrent model which can read all characters of a license plate directly. To the best of our knowledge, this is the first work that recognizes license plates without character segmentation. The whole license plate is firstly converted into a sequential feature without pre-segmentation. Here We concatenate different levels of features extracted from the $9$-layer CNN model, which combines both the local and global information.
    A bi\-di\-rectional recurrent neural network
    (BRNN) model with long short-term memory (LSTM) is employed to recognize the feature sequence. CTC was applied on the output of RNN to decode the character string in the plate. This approach takes advantage of both deep CNNs for feature learning and RNNs for sequence labelling, and results in appealing performance.

\end{itemize}

The rest of paper is organized as follows. Section $2$ gives a brief discussion on related work. Section $3$ describes the details of license plate detection, and Section $4$ presents the two methods on license plate recognition. Experimental verifications are followed in Section $5$, and conclusions drawn in Section $6$.

\section{Related work}
\label{sec:ReWork}

In this section, we would like to give a brief introduction about previous work on license plate detection and recognition. The related work on CNNs and RNNs will also be discussed shortly.

\subsubsection{License Plate Detection:} License plate detection investigates an input image and outputs potential license plate bounding boxes. Although plenty of plate detection algorithms have been proposed during the past years, it is still an challenging problem to detect license plates accurately in the wild from arbitrary capturing viewpoint, with partial occlusion or with multiple instances. The existing algorithms can be roughly classified into four categories~\cite{Du2013Automatic,Zhou2012Principal,Anagnostopoulos}: edge-based,  color-based, texture-based, and character-based.

Edge-based approaches try to find regions with higher edge density than elsewhere in the image as license plates. %
Considering the property that the brightness change in license plate region is more remarkable and more frequent than elsewhere, methods~\cite{Bai2004,Qiu2009,Zheng2005,Tan2013,Lalimi2013} use edge detector combining with some morphological operations to find the rectangles which are regarded as candidated license plate. %
In~\cite{Chen2012}, a license plate localization method based on an improved Prewitt arithmetic operator is proposed. The exact location is then determined by horizontal and vertical projections.
Rasheed~\etal~\cite{Rasheed2012} use canny operator to detect edges and then use Hough transform to find the strong vertical and horizontal lines as the bounds of the license plate. %
Edge-based methods are fast in detection speed. However, they cannot be applied to complex images as they are too sensitive to unwanted edges. It is also difficult to find license plates if they are blurry.

Color-based approaches are based on the observation that color of the license plate is usually different from that of the car body. %
HSI color model is used in~\cite{Deb2008} to detect candidate license plate regions, which are later verified by position histogram. Jia~\etal~\cite{Jia2007} firstly segment the image into different regions according to different colors via mean-shift algorithm. License plates are then distinguished based on features including rectangularity, aspect ratio and edge density.  Color-based methods can be used to detect inclined or deformed license plates. However, they cannot distinguish other objects in the image with similar color and size as the license plates. Moreover, they are very sensitive to various illumination changes in natural scene images.

Texture-based approaches try to detect license plates according to the unconventional pixel intensity distribution in plate regions. Zhang~\etal~\cite{Zhang2006} propose a license plate detection method using both global statistical features and local Haar-like features. Classifiers based on global features exclude more than $70\%$ of background area, while classifiers based on local Haar-like features are robust to brightness, color, size and position of license plates.
In~\cite{Giannoukos2006,Giannoukos2010},  sliding concentric window (SCW) algorithm is developed to identify license plate based on the local irregularity property of license plate in the texture of a image. Operator context scanning (OCS) algorithm is proposed in~\cite{Giannoukos2010} to accelerate detection speed. %
In~\cite{Yu2015}, wavelet transform is applied at first to get the horizontal and vertical details of an image. Empirical mode decomposition (EMD) analysis is then employed to deal with the projection data and locate the desired wave crest which indicates the position of a license plate. Texture-based methods use more discriminative characteristics than edge or color, but result in a higher computational complexity.

Character-based approaches regard the license plate as a string of characters and detect it via examining the presence of characters in the image. %
Lin~\etal~\cite{Lin2010} propose to detect license plate based on image saliency. This method firstly segments out characters in the image with a high recall using an intensity saliency map, then applies a sliding window on these characters to compute some saliency related features and detect license plates.
Li~\etal~\cite{Li2013}
apply maximally stable extremal region (MSER) at the first stage to extract candidate characters in images. Conditional random field (CRF) is then constructed to represent the relationship among license plate characters. License plates are finally localized through the belief propagation inference on CRF. Character-based methods are more reliable and can lead to a high recall.
However, the performance is affected largely by the general text in the image background.

Our method is a character-based approach. We use CNN to distinguish characters from cluttered background. The high classification capability of CNN model guarantees the character detection performance.
To distinguish license plates from general text in the image, another plate/non-plate CNN classifier is employed, which eliminates those hard false positives effectively.

\subsubsection{License plate recognition}
Previous work on license plate recognition typically
needs to segment characters in the license plate firstly,
and then recognizes each segmented character using optical character recognition (OCR) techniques.

Existing algorithms on license plate segmentation can mainly be
divided into two categories: projection-based and connected component (CC)-based.
Since characters and background have obviously different colors in a license plate, in theory,
they should have opposite binary values in the binary image.
Projection-based approaches exploit the histograms of vertical and horizontal pixel projections for character segmentation~\cite{Nomura2005,Guo2008,Qiao2010}. %
The binary license plate image is projected horizontally to determine the top and bottom boundaries of the characters, and then vertically to separate each character.
This type of methods can be easily affected by the rotation of license plate.
CC based methods label connected pixels in the binary license plate into components based on $4$ or $8$ neighborhood connectivity~\cite{Chang2004,Giannoukos2006,Giannoukos2010,Jiao2009}. CC based methods can extract characters from rotated license plate, but cannot segment characters correctly if they are joined together or broken apart. %
Zheng\etal~\cite{Zheng2013An} combine both methods to enhance the segmentation accuracy.
It is worth noting that both methods are implemented on binary images. Hence binarization has a significant influence on the segmentation performance. %

Apart from these methods, Zhang\etal~\cite{Zhang2013} present a character segmentation method based on character contour and template matching. Capar and Gokmen~\cite{Capar2006} propose a method that integrates the statistical boundary shape models into fast matching representation for both character segmentation and recognition. %

License plate character recognition is a kind of image classification task, where the segmented characters need to be classified into $36$ classes ($26$ upper-case letters and $10$ digits). The existing algorithms consist of template matching based and learning based methods.

Template matching based methods recognize each character by measuring the similarity between character and templates. The most similar template is regarded as the target~\cite{Rasheed2012,Goel2013,Ko2003}. Several similarity measuring methods are proposed, like Mahalanobis distance, Hausdorff distance, Hamming distance, \etc~\cite{Du2013Automatic}. Template matching methods are simple and straightforward. They are usually implemented on binary images which eliminate the change of lighting, and are used to recognize characters with single font, fixed size, no rotation and broken.

Learning based methods are often more robust. They learn information from more discriminate features, such as image density
\cite{Jiao2009,Giannoukos2010}, gradient \cite{Llorens2005}, direction features \cite{Wen2011}, LBP features
\cite{Liu2010}, and so on, and can deal with characters of different font, illumination, or rotation.
The common used learning methods include support vector machines \cite{Wen2011}, artificial neural networks
\cite{Jiao2009}, PNN \cite{Giannoukos2006,Giannoukos2010}, hidden Markov model (HMM)
\cite{Llorens2005}, \etc. Some researchers integrate multiple features
\cite{Wen2011}, or combine multiple classifiers \cite{Sharma2014} to improve the recognition accuracy.

Our first method on license plate recognition is based on this general procedure, where a CC based method is applied for character segmentation, and another deeper CNN model with much higher discriminative capability is used for character recognition.

\subsubsection{CNN and RNN techniques}

CNNs and RNNs have been applied to a number of tasks in computer vision, with breakthrough improvements achieved in recent years. Here we only introduce the ones that are related to our work. In~\cite{Wang2012}, a $3$-layer CNN model trained with unsupervised learning algorithm is developed for character classification. A $2$-way text/non-text CNN classifier is used for text detection. Another $62$-way case-sensitive character CNN classifier is employed for word recognition, combining with Viterbi-style algorithm and beam search technique.  Jaderberg~\etal~\cite{Max2014ECCV} develop a $4$-layer CNN for text spotting in natural scene. With maxout and dropout techniques, this CNN model produces much better character classification performance than that in~\cite{Wang2012}. A text/non-text CNN classifier is used in a sliding window fashion across the image for text detection. Text recognition is realized using Viterbi scoring function with character probability maps calculated by character and bigram CNNs as cues. They also publish a large character dataset with about $163$k $24 \times 24$ pixel characters cropped from natural scene images.

RNNs are powerful connectionist models which can learn time dynamics via cycles. They have been widely used to deal with sequential problems such as speech recognition~\cite{Graves2006ICML}, language translation~\cite{Sutskever2014} and handwriting recognition~\cite{Graves2009Pami}. In~\cite{Graves2009Pami}, RNN is firstly used to recognize unconstrained handwriting text without segmentation. $9$ geometrical features are computed from each sub-window across the text line. A single RNN is trained to label the sequential features. CTC is applied to the output layer of RNN to get the output labels. The authors also discuss the difference between RNNs and HMM-based methods, and demonstrate the superiority of RNN+CTC scheme. This scheme is then extended to solve the word recognition problem in~\cite{Su2014ACCV,He2015Reading}, with different sequential features adopted. To be specific, \cite{Su2014ACCV} uses histogram of oriented gradient (HoG) features, while \cite{He2015Reading} extracts features from the fully connected layer of CNN model. RNN+CTC scheme was followed for sequential feature labelling and decoding.

Our second method on license plate recognition is also motivated by this scheme. The difference is that we use a different CNN model and collect both global and local features. Experiments show that this scheme can produce promising recognition results, without any language model or lexicon provided.

\section{Car License Plate Detection}
\label{sec:Detection}

\begin{figure*}[tb]
\centering
\includegraphics[width=0.95\textwidth]{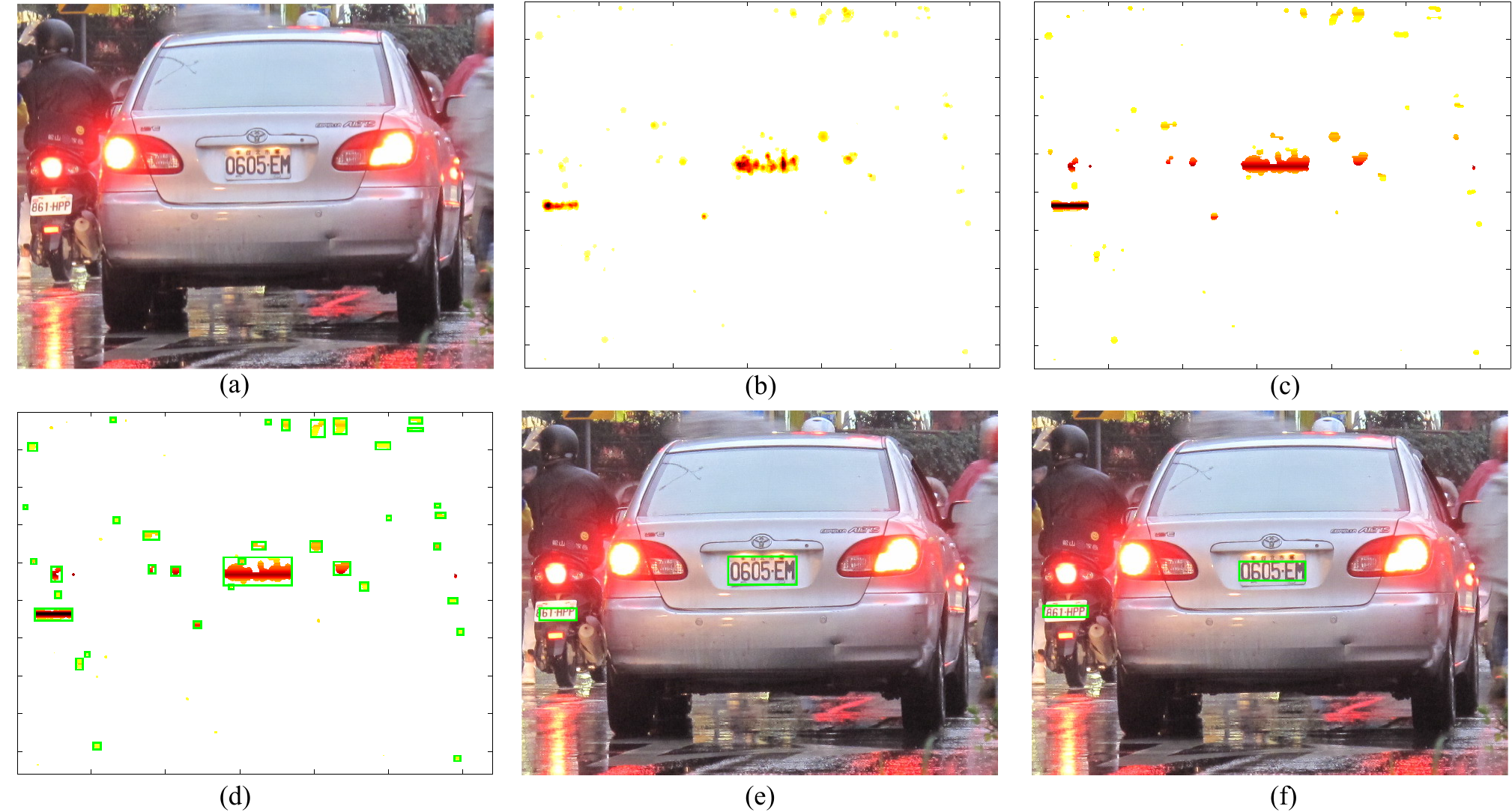}
\caption{license plate detection procedure in a single scale. (a) input image. (b) text salience map generated by CNN classifier. (c) text salience map after NMS and RLSA. (d) candidate bounding boxes generated by CCA. (e) candidate bounding boxes after false positive elimination. (f) final bounding boxes after box refinement.}
\label{Fig:2}
\end{figure*}

License plate detection is the first stage of LPDR pipeline. It has a great influence on the performance of the whole system. Ideally it is required to generate plate bounding boxes with high recall and high precision. In this work, we take advantage of the high discriminant ability of CNNs and perform a character-based plate detection using a multiple scale sliding window manner. Firstly, we train CNN classifier based on characters cropped from general text. One benefit that comes along with this method is that our classifier can be used to detect plate from different countries, as long as the plate is composed of digitals and upper-case letters. %
In order to distinguish license plate from other text or text-like outliers appeared in the image, %
another plate/non-plate CNN classifier is employ to remove those false positives. We also refine the bounding boxes based on projection based method to improve the overlap ratio.  The overall detection process is illustrated in Fig.~\ref{Fig:2}.

\setcounter{subsubsection}{0}
\subsubsection{Candidate License Plate Detection}
In order to accelerate the detection process, we train a $4$-layer CNN to classify whether the image patch contains characters or not. Due to the convolutional structure of CNN, we compute the character saliency map by running CNN classifier across the entire image in one go, instead of calculating on each cropped sliding window, which can save processing time.

The configuration of $4$-layer character CNN model is shown in Table~\ref{Tab:1}. It includes $2$ convolutional layers and $2$ fully connected layers. The last $37$ channels are fed into a soft-max layer to convert them into prediction probabilities. %
Here we train a $37$-class CNN classifier for $26$ upper-case letters, $10$ digitals and a non-character class, instead of a binary text/non-text classifier. Patches classified as either letters or digitals are all regarded as characters later. With this way the features learned for each class are more specific and discriminative, and will lead to a better detection result with a higher recall.

\begin{table} [ht]
	\begin{center}
	\caption{Configuration of the $4$-layer Character CNN model}
	\label{Tab:1}
	{
	\begin{tabular}{c|c}
	\hline
	 Layer Type & Parameters  \\
	\hline
	Soft-max & $37$ classes \\
	\hline
	Fully connected & \#neurons: $37$ \\
	\hline
	Dropout & Prop: $0.5$ \\
	\hline
	ReLU & \\
	\hline
	Fully connected & \#neurons: $512$ \\
 	\hline
 	Maxpooling & p: $2 \times 2$, s: $2$ \\
   \hline
    ReLU & \\
 	\hline
 	Convolution & \#filters: $384$, k:$2 \times 2$, s:$1$, p:$0$ \\
	\hline
	Maxpooling & p: $4 \times 4$, s: $4$ \\
	\hline
	ReLU & \\
	\hline
	Convolution & \#filters: $120$, k:$5 \times 5$, s:$1$, p:$0$ \\
	\hline
	Input & $24 \times 24$ pixels gray-scale image \\
	\hline
	\end{tabular}
	}

	\end{center}
\end{table}

CNN is trained with grayscale characters and background images of $24$ $\times$ $24$ pixels, normalized by substracting the mean over all training data. Data augmentation is carried out by random image translations and rotations to reduce overfitting. We train CNN using stochastic gradient descent (SGD)
with back propagation. An adaptively reduced learning rate is adopted to ensure a fast and stable training process. %
Bootstrapping is also employed to deal with the cluttered background information, \ie, wrongly classified negative examples are
collected to retrain the CNN model. %

For license plate detection, the first phase is to generate candidate license plate bounding boxes with a high recall. Given an input image, we resize it into $12$ different scales, and calculate the character saliency map at each scale by evaluating CNN classifier in a sliding window fashion across the image.  The input image is padded with $12$-pixels each side so that characters near image edges would not be missed. After getting these saliency maps, the character string bounding boxes are generated independently at each scale by using the run length smoothing algorithm (RLSA)~\cite{Max2014ECCV} and connected component analysis (CCA). In detail, for each row in the saliency map, we do non-maximal suppression (NMS) at first to remove detection noise. NMS response for the pixel located at row $r$ column $x$ with classification probability $P(x,r)$ is defined as follows:

\begin{equation}
\hat{P}(x,r) =
\begin{cases}
P(x,r) & if \,\, P(x,r) \ge P(x',r), \\
       & \forall x' \,\, \sst \,\, \|x' - x\| < \delta \\
0 & Otherwise
\end{cases}
\end{equation}
where $\delta$ defines a width threshold.
Then we calculate the mean and standard deviation of the spacings between probability peaks. Neighboring pixels are connected together if the spacing between them is less than a threshold. CCA is applied subsequently to produce the initial candidate boxes. The process is shown in Fig.~\ref{Fig:2}. %

\subsubsection{False Positives Elimination and Bounding Box Refinement}
After all the scaled images are processed, the produced bounding boxes are firstly filtered based on some geometric constraints (boxes length, height, aspect ratio, \etc). Then we score each box by averaging the character saliency score within it. Boxes whose scores are less than the average box score are eliminated. NMS is employed again on the bounding box level.

\begin{figure}[tb]
\centering
\includegraphics[width=0.3\textwidth]{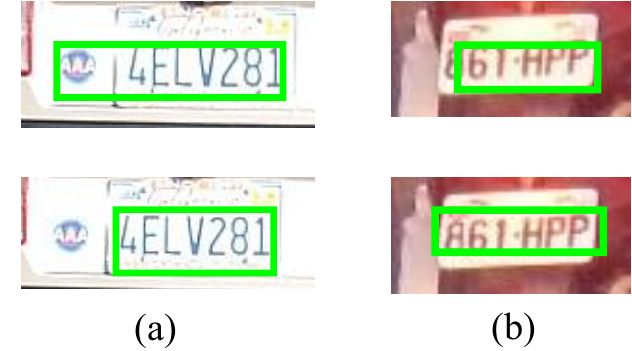}
\caption{Generated bounding boxes before and after bounding box refinement. The top line shows the initial bounding boxes, and the bottom line is the results after refinement. }
\label{Fig:41}
\end{figure}

We find that some generated bounding boxes are too big or too small, which will affect the following plate verification and recognition. For example, the textual-like background contained in the bounding box in Fig.~\ref{Fig:41}(a) will impact the following character segmentation. The bounding box in Fig.~\ref{Fig:41}(b) that does not contain the whole license plate will definitely lead to a incorrect recognition result. Therefore, a process for refining bounding boxes is performed according to the edge feature of license plate~\cite{Zheng2013An}.

\begin{figure}[tb]
\centering
\includegraphics[width=0.4\textwidth]{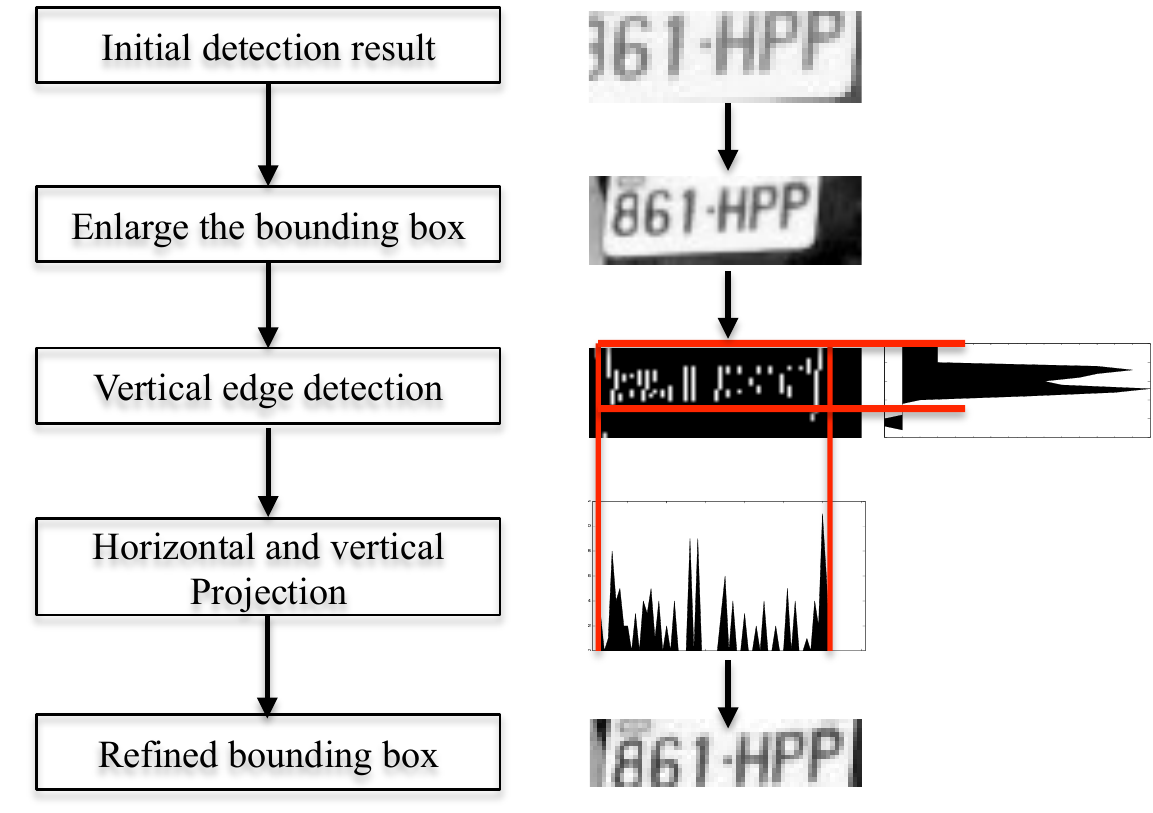}
\caption{The process of bounding box refine.}
\label{Fig:5}
\end{figure}

For each detected bounding box, we enlarge the box with $20\%$ on each side. Considering the strong connectivity of characters in vertical direction than in horizontal direction, we perform vertical edge detection on the cropped license plate image using Sobel operator. When we get the vertical edge map, a horizontal projection is performed to find the top and bottom boundaries of the license plate. Then a vertical projection is carried out to get the left and right bounds of the license plate. The process is presented in Fig.~\ref{Fig:5}.

Finally we use another plate/non-plate CNN classifier to verify the remaining bounding boxes. The binary plate/non-plate CNN model is presented in Table~\ref{Tab:3}.  It is trained with positive samples of grayscale license plates from different countries, either cropped from real images or synthesized by ourselves, and negative samples constituted by non-text image patches as well as some general text strings. The size of the input image is $100 \times 30$ pixels. Data augmentation and bootstrapping are also applied here to improve the classification performance. For each candidate license plate, we evaluate it by averaging the probability of five predictions over random image translation, so as to remove noises. The ones that are classified as license plates are fed to the next step.

\begin{table} [ht]
	\begin{center}
	\caption{Configuration of the $4$-layer license plate/Non-plate CNN model}
	\label{Tab:3}
	{
	\begin{tabular}{c|c}
	\hline
	 Layer Type & Parameters  \\
	\hline
	Soft-max & $2$ classes \\
	\hline
	Fully connected & \#neurons: $2$ \\
	\hline
	Dropout & Prop: $0.5$ \\
	\hline
	ReLU & \\
	\hline
	Fully connected & \#neurons: $500$ \\
 	\hline
 	Maxpooling & p: $3 \times 3$, s: $3$ \\
   \hline
    ReLU & \\
 	\hline
 	Convolution & \#filters: $256$, k:$5 \times 5$, s:$1$, p:$0$ \\
	\hline
	Maxpooling & p: $2 \times 2$, s: $2$ \\
	\hline
	ReLU & \\
	\hline
	Convolution & \#filters: $96$, k:$5 \times 5$, s:$1$, p:$0$ \\
	\hline
	Input & $30 \times 100$ pixels gray-scale image \\
	\hline
	\end{tabular}
	}

	\end{center}
\end{table}

\section{Car License Plate Recognition}
\label{sec:Recognition}

The second stage of LPDR system is to recognize the characters in the plate. Although there are commercial softwares available for optical character recognition (OCR), they are mainly for scanned documents, and cannot be applied directly to the natural images due to variations in illumination, distortion and viewpoint rotation. In this work, we present two different methods on plate recognition. The first one serves as a baseline. It performs character segmentation firstly and then recognizes each character.  The other one regards the character string recognition as a sequence labelling problem, and recognizes all characters in the license plate one-off.  This method avoids the challenging task of character segmentation. However, it needs a large number of labelled license plates as training instances, which may limit its application. Next we will introduce both methods one by one.

\subsection{Character Segmentation Based Plate Recognition}
\label{sec:Segmentation}
As demonstrated  in Fig.~\ref{Fig:1}, there are mainly three phases in the character segmentation based plate recognition approach.

\subsubsection{Image Binarization}
The first phase performs image binarization. When the candidate license plate is cropped from the original image, an intensity adjustment is applied to the grayscale image such that $1\%$ of pixels are saturated at low and high intensities of original image. This step helps to increases the image contrast, and is beneficial for the subsequent image binarization. %

There are several methods for image binarization. Global thresholding methods, like Otsu's method, use a single threshold to segment foreground regions, which are easy to make mistakes under unbalanced illumination conditions. Therefore we prefer local thresholding methods here. We evaluated several local binarization methods, including Niblack's, Sauvola's, Howe's, \etc~\cite{Milyaev2015}, and found that Non-linear Niblack's algorithm produced much better results in our case. Non-linear Niblack' algorithm adds two ordered statistics filter to the background and foreground filters respectively, which can effectively handle poor conditions in natural scenes, such as uneven illumination and degraded text. %

After performing image binarization, there is a key step that we need to determine which pixels in the binary image represent characters, the black ones or the white ones. Because CCA used later generates blocks based on white pixels, the binary images should use white pixels as foreground and dark pixels as background. Here we adopt a simple idea which assumes that background area in the license plate is generally larger than the text area. If the number of white pixels as considered as characters is much more than the blacks, the binary image needs to be inverted.

\subsubsection{Connected Component Analysis}
CCA is adopted here on the binary image which links the connected ingredients together as character blocks. %
CCA is simple to perform and is robust to license plate rotation. Nevertheless, it can result in a lot of false positives, such as the plate edge, subtitle, irrelevant marks or bars. Sometimes several character may be connected together into one block, whereas sometimes one character maybe separated into two blocks because of the uneven lighting or shadow. So we remove those false blocks, and perform block splitting and merging based on some geometrical rules. The details refer to \cite{Yoon2011}.

\subsubsection{Character Recognition}
The last phase is to recognize the candidate characters segmented from the license plate in the previous step.
Here we design another deeper CNN model with $9$ layers for $36$-way character recognition, as shown in Table~\ref{Tab:2}. Non-character class is not involved in this case so as to exclude the influence of background information and improve the recognition accuracy on characters. The CNN model includes $6$ convolutional layers and $3$ fully connected layers. A soft-max layer is followed lastly to provide the prediction probabilities on each class.
We use the same training dataset as that used in the previous $4$-layer $37$-class CNN classifier. More data augmentations are implemented including random rotations, translations and noise injection. Those data is normalized by substracting the mean over all training data. Dropout is involved in every fully connected layers with a proportion of $0.5$ to prevent over-fitting. This complicated CNN structure will learn more sophisticated features and lead to a higher recognition accuracy. We still use SGD with BP to train the CNN classifier. The reason why we do not use this complicated CNN model during license plate detection is due to the concern of computation complexity. $4$-layer CNN would be much faster with a sufficiently high recall, whereas false positives can be rejected effectively by the subsequent process.

\begin{table} [ht]
	\begin{center}
	\caption{Configuration of the $9$-layer CNN model}
	\label{Tab:2}
	{
	\begin{tabular}{c|c}
	\hline
	 Layer Type & Parameters  \\
	\hline
	Soft-max & $36$ classes \\
	\hline
	Fully connected & \#neurons: $36$ \\
	\hline
	Dropout & Prop: $0.5$ \\
	\hline
	ReLU & \\
	\hline
	Fully connected & \#neurons: $1000$ \\
	\hline
	Dropout & Prop: $0.5$ \\
	\hline
	ReLU & \\
	\hline
	Fully connected & \#neurons: $1000$ \\
	\hline
	Maxpooling & p: $3 \times 3$, s: $1$ \\
	\hline
    ReLU & \\
 	\hline
 	Convolution & \#filters: $512$, k:$3 \times 3$, s:$1$, p:$1$ \\
	\hline
    ReLU & \\
 	\hline
 	Convolution & \#filters: $512$, k:$3 \times 3$, s:$1$, p:$1$ \\
	\hline
	Maxpooling & p: $3 \times 3$, s: $2$ \\
	\hline
    ReLU & \\
 	\hline
 	Convolution & \#filters: $256$, k:$3 \times 3$, s:$1$, p:$1$ \\
	\hline
    ReLU & \\
 	\hline
 	Convolution & \#filters: $256$, k:$3 \times 3$, s:$1$, p:$1$ \\
 	\hline
 	Maxpooling & p: $3 \times 3$, s: $2$ \\
   \hline
    ReLU & \\
 	\hline
 	Convolution & \#filters: $128$, k:$3 \times 3$, s:$1$, p:$1$ \\
	\hline
	Maxpooling & p: $3 \times 3$, s: $1$ \\
	\hline
	ReLU & \\
	\hline
	Convolution & \#filters: $64$, k:$3 \times 3$, s:$1$, p:$1$ \\
	\hline
	Input & $24 \times 24$ pixels gray-scale image \\
	\hline
	\end{tabular}
	}

	\end{center}
\end{table}

Rich and discriminative features play an important role in the classification performance. Inspired by~\cite{Zhong2015High} which adds directional feature maps (Gabor, Gradient and HoG) into the input of CNN, and shows improved performance on handwritten Chinese character recognition, in this work, we incorporate local binary pattern (LBP) features into CNN. LBP is a powerful texture spectrum model which is robust against illumination variation and fast to compute. It shows promising results for character recognition~\cite{Hsu2013,Chen2011LBP}.
We extract LBP features for each pixel in the image patch from a $3 \times 3$ pixel neighborhood. The produced feature map is concatenated to the original grayscale image and formed a $24 \times 24 \times 2$ array as the CNN input layer. Then we fine-tune the $9$-layer CNN model with the additional features. %

The segmented candidate character is cropped from the license plate, and resized into $24 \times 24$ pixels for recognition. In order to improve recognition accuracy, we augment the test data as well by randomly cropping five patches from the neighborhood of segmented character. The final recognition result is based on the average of five prediction probabilities produced by the network's soft-max layer. We test the performance of each classifier respectively, and also combine the predictions of the two classifiers as the final result. The recognition results are presented in the experiments.

The number of characters in the license plate usually has a restriction for each country, for example, no more than $N$. If there are more than $N$ segmented blocks in the previous phase, we will reject the ones with lower prediction probabilities, and return a recognition result with $N$ characters.

\subsection{Sequence Labelling based Plate Recognition}
\label{sec:Sequence}
In the traditional framework of LPDR, as we introduced above, character segmentation has a great influence on the success of plate recognition. The license plate will be recognized incorrectly if the segmentation is improper, even if we have a recognizer with high performance which can deal with characters of various size, font and rotation. However, the character segmentation process by itself is a really challenging task that is prone to be influenced by uneven lighting, shadow and noise in the image. A lot of rules are usually employed in previous work~\cite{Yoon2011,Zheng2013An,Hsu2013} to modify the improperly segmented blocks, which are not robust. In this part, we use a novel recognition technique that treats the characters in license plate as an unsegmented sequence, and solve the problem from the viewpoint of sequence labelling.

\subsubsection{Overview}
\begin{figure}[tb]
\centering
\includegraphics[width=0.45\textwidth]{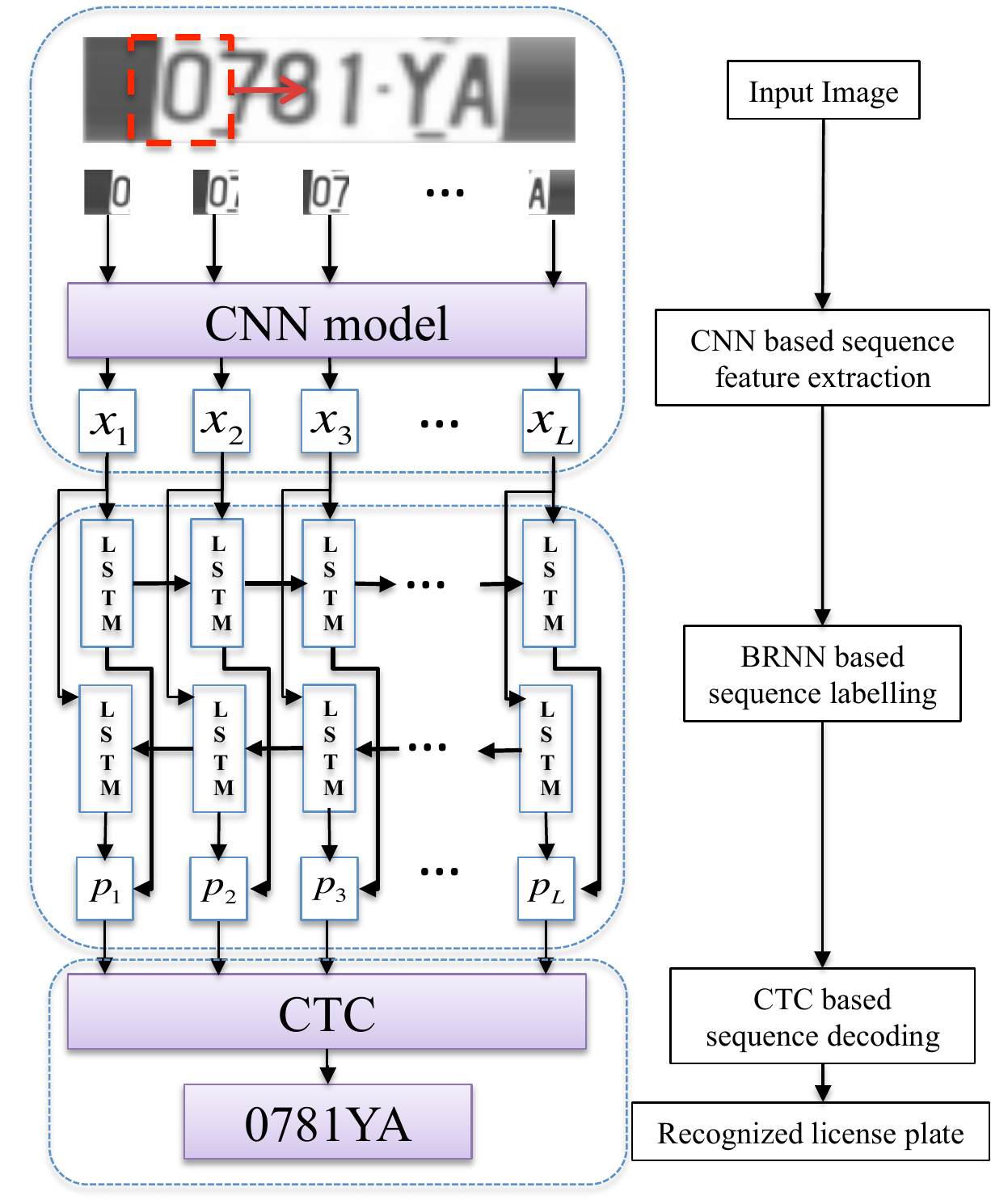}
\caption{Overall structure of the sequence labelling based plate recognition.}
\label{Fig:3}
\end{figure}

The overall procedure of our sequence labelling based plate recognition is presented in Fig.~\ref{Fig:3} . It mainly consists of three sub-parts.

To begin with, the license plate bounding box is converted into a sequence of feature vectors which are extracted by using the pre-trained $9$-layer CNN model sliding across the bounding box.
Then a bidirectional RNN model with LSTM is trained to label the sequential features, with SGD and BP algorithm. CTC is applied lastly to the output layer of RNN, which analyzes the labelling results of RNN and generates the final recognition result. This method allows us to recognize the whole license plate without character level segmentation and recognition. The bidirectional RNN can help to capture the interdependencies between features in the sequence and works reliable for degraded images. The high level feature learned by our sophisticated CNN model is also crucial to the success of this method.

\subsubsection{Sequence Feature Generation}
CNN has demonstrated strong capabilities of learning informative representations from image. Here the pretrained $9$-layer CNN model is applied to extract a sequential feature representation from the cropped license plate image. Inspired by the work of~\cite{Su2014ACCV,He2015Reading}, the features are extracted in a sliding window manner across the image.

Firstly, we need some preprocessing. For each detected license plate, we convert it to grayscale, and padded with $10$ pixels on both left and right sides so as to retain edge information of the plate. Then we resize the plate to $24 \times 94$ pixels, with the height the same as the input height for our CNN model. %

After that we use a sub-window of size $24 \times 24$ pixels to partition the padded image convolutionally, with a step size of $1$.  %
For each partitioned image patch, we feed it into the $9$-layer CNN model,  and extract the $4 \times 4 \times 256$ feature vector from the output of the fourth convolutional layer, as well as the $1000$ feature vector from the output of the first fully connected layer. The two vectors are then concatenated together into one feature vector with length $5096$.  This feature contains both local and global information of the image patch, which will bring better recognition performance compared with the one extracted only from the fully connected layer. PCA is applied to reduce the feature dimension to $256$D, followed by a feature normalization.

With this operation, features are extracted from left to right on each sub-window in the candidate license plate image, and form a feature sequence array $\textbf{x} = \{x_1,x_2,\dots,x_L\}$, where $x_t \in \mathcal{R}^{256}$, $L$ is the width of the resized license plate.
It not only keeps the order information, but also captures sufficient context information for RNN to exploit. The inter-relation between feature vectors contributes a lot to character recognition.

\subsubsection{Sequence Labelling}
RNN is a special neural network which provides a powerful mechanism to exploit past contextual information.
These contextual cues will make the sequence recognition more stable than treating each feature independently.  To overcome the shortcoming of gradient vanishing or exploding during RNN training, LSTM is employed. It contains a memory cell and three multiplicative gates, which can store the contexts for a long period of time, and capture long-range dependencies between sequential features.

For our task of character string recognition, it would be helpful to have access to the context both in the past and in the future. So bidirectional-RNNs (BRNNs) are applied here. As illustrated in Figure , there are two separated hidden layers in BRNNs, one of which processes the feature sequence forward, while the other one processes it backward. For each hidden layer, all LSTMs share the same parameters. Both hidden layers are connected to the same output layer, providing it with information in both directions along the input sequence.

Sequence labelling is processed by recurrently implementing BLSTMs for each feature in the feature sequence. Each time they update the state $h_{t}$ with a non-linear function $g(\cdot)$ that takes both current feature $x_t$ and neighboring state $h_{t-1}$ or $h_{t+1}$ as inputs, \ie,
\begin{equation}
\begin{cases}
h_t^{(f)} = g(x_t, h_{t-1}^{(f)}), \\
h_t^{(b)} = g(x_t, h_{t+1}^{(b)}).
\end{cases}
\end{equation}
A soft-max layer follows which transforms the LSTMs' states into a probability distribution over $37$ classes, \ie,
\begin{equation}
p_t(c=c_k | x_t) = softmax(h_t^{(f)},h_t^{(b)}), \,\, k=1,\dots,37
\end{equation}
The whole feature sequence is finally transformed into a sequence of probability estimation $\textbf{p} = \{p_1,p_2,\dots,p_L\}$ with the same length as the input sequence.

\subsubsection{Sequence Decoding}

Last, we need to transform the sequence of probability estimation $\textbf{p}$ into character string. We follow the handwriting recognition system~\cite{Graves2009Pami} by applying a CTC to the output layer of RNN. CTC is specifically designed for sequence labelling tasks without the need of data pre-segmentation. It uses a network to decode the prediction sequence into output labels directly.

The objective function of CTC is defined as the negative log probability of the network correctly labelling the entire training set, \ie

\begin{equation}
\mathcal{O}=-\sum_{(\textbf{c},\textbf{z}) \in \mathcal{S}} \ln P(\textbf{z} | \textbf{c}),
\end{equation}
where $\mathcal{S}$ is the taring dataset, which consists of pairs of input and target sequences $(\textbf{c}, \textbf{z})$. $P(\textbf{z} | \textbf{c})$ denotes the conditional probability of obtaining target sequence $\textbf{z}$ through the input $\textbf{c}$. The target is to minimize $\mathcal{O}$, which is equivalent to maximize $P(\textbf{z} | \textbf{c})$.

This network can be trained with gradient descent with back propagation.  We connect it directly to the outputs of BRNNs, which can be regarded as a kind of loss function. So the input of CTC $\textbf{c}$ is exactly the output activations of BRNNs \textbf{p},
and the aim of sequence decoding is to find an approximately optimal path $\pi$ with maximum probability through the LSTMs output sequence, \ie,
\begin{equation}
\textbf{l}^* \thickapprox \mathcal{B}(\arg \max_{\pi} P(\pi | \textbf{p})).
\end{equation}
Here a path $\pi$ is a label sequence based on the output activations of BRNNs:
\begin{equation}
P(\pi | \textbf{p}) = \prod_{t=1}^{L}P(\pi_t|\textbf{p}) =\prod_{t=1}^{L} p_t(c=\pi_t|x_t).
\end{equation}
An operator $\mathcal{B}$ is defined which removes the repeated labels and the non-character labels from the path. For example, $\mathcal{B}(a-a-b-)=\mathcal{B}(-aa--ab-b)=(aab)$.
Details of CTC can refer to~\cite{Graves2006ICML,Graves2009Pami}.

\section{Experiments}
\label{SEC:Exp}
In this section, experiments are performed to verify the effectiveness of the proposed methods. %
Our experiments are implemented on NVIDIA Tesla K$40$c GPU with $6$GB memory.
The CNN models are trained using MatConvNet~\cite{vedaldi15matconvnet}.

\subsection{Dataset}
A sufficiently large training dataset is essential for the success of a CNN classifier. The $4$-layer character CNN classifiers is trained on roughly $1.38 \times 10^5$ character images and $9 \times 10^5$ non-character images, and the $9$-layer CNN model is trained only with the character images.  The character images comprise $26$ upper-case letters and $10$ digitals sampled from the dataset created by Jaderberg~\etal~\cite{Max2014ECCV} and Wang \etal~\cite{Wangkai2011}. The non-character image patches are cropped by ourselves from the ICDAR dataset~\cite{icdar2003,icdar2005,icdar2011} and Microsoft Research Cambridge Object Recognition Image database \cite{msrcorid}. All images are in grayscale and resized to $24 \times 24$ pixels for training. Data augmentation is carried out by image translations and rotations to reduce overfitting. Bootstrapping, which collects hard negative examples and re-trains the classifier, is also employed to improve the classification accuracy.

As to the license plate/non-plate dataset. We cropped around $3000$ license plates images from public available dataset~\cite{Zhou2012Principal,Cnagnost2008}. We also synthesize nearly $5 \times 10^4$ license plates using ImageMagic, following the fonts, colors and composition rules of real plates, adding some amount of Gaussian noise and applying a random lighting and affine deformation. Around $4 \times 10^5$ background images are used here including patches without any characters and patches with some general text. All the images are grayscale and resized to $100 \times 30$ pixels for training. Date augmentation and bootstrapping are also adopted to improve performance.

We test the effectiveness of the proposed detection and recognition algorithms on two datasets. The first one is the Caltech Cars (Real) $1999$ dataset~\cite{carmarkus} which consists of $126$ images with resolution of $896 \times 592$ pixels. The images are taken in the Caltech parking lots, which contains a USA license plate with cluttered background, such as trees, grass, wall, \etc. The second dataset is the application-oriented license plate (AOLP) benchmark database~\cite{Hsu2013}, which has $2049$ images of Taiwan license plates. This database is categorized into three subsets: access control (AC) with $681$ samples, traffic law enforcement (LE) with $757$ samples, and road patrol (RP) with $611$ samples. AC refers to the cases that a vehicle passes a fixed passage with a lower speed or full stop. This is the easiest situation. The images are captured under different illuminations and different weather conditions. LE refers to the cases that a vehicle violates traffic laws and is captured by roadside camera. The background are really cluttered, with road sign and multiple plates in one image. RP refers to the cases that the camera is held on a patrolling vehicle, and the images are taken with arbitrary viewpoints and distances. The detailed introduction of this AOLP dataset can be found in~\cite{Hsu2013}.

\subsection{Evaluation Criterion}
As stated in~\cite{Du2013Automatic}, there is no uniform way to evaluate the performance of different LPDR systems. In this work, we follow the evaluation criterion for general text detection in natural scene, and quantify the detection results using precision/recall rate~\cite{icdar2003}. Precision is defined as the number of correctly detected license plates divided by the total number of detected regions. It gives us information on the amount of false alarms. Systems that over-estimate the number of bounding boxes are punished with a low precision score. Recall is defined as the number of correctly detected license plates divided by the total number of groundtruth. It measures how many groundtruth objects have been detected. Systems that under-estimate the number of groundtruth are punished with a low recall score. Here a detection is considered to be correct if the license plate is totally encompassed by the bounding box, and the overlap between the detection and groundtruth bounding box is greater than $0.5$, where the overlap means the area of intersection divided by the area of the minimum bounding box containing both rectangles (IoU).

As to license plate recognition, we evaluate it with recognition accuracy, which is defined as the number of correctly recognized license plates divides by the total number of groundtruth. A correctly recognized license plate means all the characters on the plate are recognized correctky. In order to compare with previous work, we also give out the character recognition accuracy, which is defined as the number of correctly recognized characters divided by the total number of characters from the groundtruth. The license plates for recognition is from the detection result, rather than cropped directly from the groundtruth. Therefore, the detection performance greatly affects the final recognition result, not only quantity, but also quality.

\subsection{Character Classification Performance}
In this work, we designed several CNN models for character classification which are used under different conditions: a $4$-layer CNN model for fast detection, a $9$-layer CNN model for accurate recognition, and another $9$-layer CNN model with LBP features for further improvement. Here, we present the classification performance of these CNN models, and also compare with previous work.

As we introduced before, Jaderberg~\etal~\cite{Max2014ECCV} developed a $4$-layer CNN model for text spotting in natural scene. In this CNN model, there are nearly $2.6$M parameters. In order to accelerate detection process, we design another $4$-layer CNN model (as presented in Table~\ref{Tab:1}), which only has $1$M parameters. However, the classification performance has not been affected significantly. We train both CNN models for $37$-way classification, using the training data introduced above. The classification accuracy is evaluated via validation data, which consists of $2979$ test data from~\cite{Max2014ECCV}, excluding lower-case letters, and $3000$ non-character images cropped by ourselves. The experimental results in Table~\ref{Tab:4} show that our CNN model gives a comparable classification accuracy as Jaderberg~\etal's CNN. However, our CNN uses less parameters and is about $3-5$ times faster in both training and testing.

\begin{table} [ht]
	\begin{center}
	\caption{Classification performance of different CNN models on $37$-way characters ($26$ upper-class letters, $10$ digits and non-character). Jaderberg~\etal's CNN model is a little bit better than ours, but has $2$ times more parameters, which will result in longer training and detection process.}
	\label{Tab:4}
	{
	\begin{tabular}{l|c|c}
	\hline
	 CNN Model & \#Parameters & Classification Error  \\
	\hline
	 Jaderberg~\etal~\cite{Max2014ECCV} & $2.6$M & $\textbf{0.0561}$ \\
	\hline
     Ours & $\textbf{1M}$ & $0.0592$ \\
	\hline
	\end{tabular}
	}
	\end{center}
\end{table}

The $36$-class CNN classifiers are tested with only the $2979$ character test data. Jaderberg~\etal's CNN model is also retrained without non-character images for fare comparison. The results in Table~\ref{Tab:5} show that our $9$-layer CNN model gives much better classification performance. Incorporating LBP features can improve the performance of CNN furthermore.
To distinguish those CNNs, we denote the $9$-layer CNN model trained with only grayscale image as "CNN I", the $9$-layer CNN model trained with LBP features included as "CNN II".
Actually we also tested the classification performance of CNN by incorporating Gabor features as demonstrated in~\cite{Zhong2015High}. The comparison result shows that using LBP features results in lower classification error.

\begin{table*} [ht]
	\begin{center}
	\caption{Classification performance of different CNN on $36$-way characters ($26$ upper-class letters and $10$ digits). Our $9$-layer CNN model gives much better classification result than Jaderberg~\etal's CNN. The performance can be further enhanced with LBP features as input.}
	\label{Tab:5}
	{
	\begin{tabular}{l|c|c|c|c}
	\hline
	 Method & Jaderberg~\etal's CNN~\cite{Max2014ECCV} & $9$-layer CNN I  & $9$-layer CNN II & $9$-layer CNN with Gabor  \\
	\hline
	 Classification Error & $0.0865$  & $0.0614$ & $\textbf{0.0580}$ & $0.0608$  \\
	\hline
	\end{tabular}
	}

	\end{center}
\end{table*}

\subsection{License Plate Detection}
The detection performance of our cascade CNN based method is shown in Table~\ref{Tab:6} for Caltech cars dataset~\cite{carmarkus}, and in Table~\ref{Tab:7} for AOLP dataset~\cite{Hsu2013}. Some previous approaches are also evaluated for comparison. The work of Le \& Li~\cite{Le2006} and Bai \& Liu~\cite{Bai2004} are edge-based methods, where color information is integrated in~\cite{Le2006} to remove false positives. Lim \& Tay~\cite{Wooi2010} and Zhou~\etal~\cite{Zhou2012Principal}'s work are character-based methods. In particular, Lim \& Tay~\cite{Wooi2010} used MSER to detect characters in the images. SIFT-based unigram classifier was trained to removed false alarms. Zhou~\etal~\cite{Zhou2012Principal} discovered the principal visual word (PVW) for each character with geometric context. The license plates were extracted by matching local features with PVW. The detection approach in Hsu~\etal~\cite{Hsu2013} is also an edge-based manner, where EM algorithm was applied for the edge clustering which extracts the regions with dense sets of edges and with shapes similar to plates as the candidate license plates.

\begin{table} [ht]
\newcommand{\tabincell}[2]{\begin{tabular}{@{}#1@{}}#2\end{tabular}}
	\begin{center}
	\caption{Comparison of plate detection results by different methods on Caltech cars dataset. Our cascade CNN based method produced the best detection result, with both the highest precision and recall.}
	\label{Tab:6}
	{
	\begin{tabular}{l|c|c}
	\hline
	 Method & Precision (\%) & Recall (\%)  \\
	\hline
	Le \& Li~\cite{Le2006} & $71.40$ & $61.60$ \\
	\hline
	Bai \& Liu~\cite{Bai2004} & $74.10$ & $68.70$ \\
	\hline
	Lim \& Tay~\cite{Wooi2010} & $83.73$ & $90.47$ \\
	\hline
	Zhou \etal~\cite{Zhou2012Principal} & $95.50$ & $84.80$  \\
	\hline
	\tabincell{l} {Ours \\ (with $37$-way outputs)}  & $\textbf{97.56}$ & $\textbf{95.24}$ \\
	\hline
	\tabincell{l} {Ours \\ (with $2$-way outputs)} & $97.39$ & $89.89$ \\
	\hline
	\end{tabular}
	}

	\end{center}
\end{table}

Based on the evaluation criterion described above, our approach outperforms all the five methods in both precision and recall on both datasets. To be specific, on Caltech cars dataset, it achieves a recall of $95.24\%$, which is $4.77\%$ higher than the second best one achieved by Lim \& Tay's method~\cite{Wooi2010}. The precision of our approach is $97.56\%$, which is also the best, with $2.06\%$ higher than the second. On AOLP dataset, our method gives the highest precisions and recalls on all three sub-datasets, with an even obvious superiority on precision. With GPU, it needs about $5$ seconds to process a image from Caltech cars dataset, and $2-3$ seconds for AOLP dataset.

The last row in Table~\ref{Tab:6} shows a detection result using our framework with $2$-way outputs. As we introduced before, in our detection phase, we use a $37$-way CNN classifier instead of a binary text/non-text classifier.
The $37$-way CNN classifier can learn the features of each character more clearly and fairly. In contrast, the $2$-way CNN classifier that put all characters in one class may omit features specific to certain characters, which is inaccurate and may miss some characters during detection. The detection result in Caltech cars dataset proves this point, where the $2$-way classifier has a lower recall compared with the $37$-way classifier.

\begin{table*} [ht]
	\begin{center}
	\caption{Comparison of plate detection results by different methods on AOLP dataset}
	\label{Tab:7}
	{

	\begin{tabular}{l|c|c|c|c|c|c}
	\hline
  \diagbox{Method}{Subset} & \multicolumn{2}{|c}{AC (\%)} & \multicolumn{2}{|c}{LE (\%)} & \multicolumn{2}{|c}{RP (\%)} \\\cline{2-7} & \multicolumn{1}{c|}{Precision} & \multicolumn{1}{|c|}{Recall}  &  \multicolumn{1}{c|}{Precision} & \multicolumn{1}{|c|}{Recall}  &  \multicolumn{1}{c|}{Precision} & \multicolumn{1}{|c}{Recall} \\
	\hline
	Hsu~\etal~\cite{Hsu2013} & $91$ & $96$ & $91$ & $95$ & $91$ & $94$  \\
	\hline
	Our approach & $98.53$ & $98.38$ & $97.75$ & $97.62$ & $95.28$ & $95.58$ \\
	\hline
	\end{tabular}
	}

	\end{center}
\end{table*}

\subsection{License Plate Recognition}
The recognition performance of the two methods are presented in Table~\ref{Tab:8}, including both datasets. Here we mainly compared with the work in~\cite{Hsu2013} as it showed higher recognition rate than some previous works such as ANN, PNN as well as the famous OCR software \emph{Tesseract}. Nevertheless, our $9$-layer CNN classifier produced even better results. In~\cite{Hsu2013}, LBP features are extracted from each character and classified using linear discriminant analysis. It only presented the character recognition results. The overall rate shown in that paper is the multiplication of the detection, segmentation and character recognition rates.
The plate recognition accuracy did not shown in that paper. So for fare comparison, we also collect all correctly recognized characters in the first approach, and calculate the character recognition rate. %
Experimental results in Table~\ref{Tab:8} show that our CNN classifier gives higher accuracy.

\begin{table*} [ht]
\newcommand{\tabincell}[2]{\begin{tabular}{@{}#1@{}}#2\end{tabular}}
	\begin{center}
	\caption{Comparison of plate detection results by different methods on AOLP dataset}
	\label{Tab:8}
	{
	\begin{tabular}{l|c|c|c|c|c|c|c|c}
	\hline
	\diagbox{Method}{Subset} & \multicolumn{2}{|c}{AC (\%)} & \multicolumn{2}{|c}{LE (\%)} & \multicolumn{2}{|c}{RP (\%)} & \multicolumn{2}{|c}{Caltech cars (\%)} \\\cline{2-9} & \multicolumn{1}{c|}{Plate} & \multicolumn{1}{|c|}{Character}  &  \multicolumn{1}{c|}{Plate} & \multicolumn{1}{|c|}{Character}  &  \multicolumn{1}{c|}{Plate} & \multicolumn{1}{|c|}{Character} &  \multicolumn{1}{c|}{Plate} & \multicolumn{1}{|c}{Character} \\
	\hline

	Hsu~\etal~\cite{Hsu2013} & $-$ & $96$ & $-$ & $94$ & $-$ & $95$ & $-$ & $-$ \\
	\hline
	\tabincell{l} {Our $1$st approach \\ (with CNN I)} & $93.53$ & $97.84$ & $89.83$ & $97.27$ & $86.58$ & $95.57$ & $82.54$ & $90.48$ \\
	\hline
	\tabincell{l} {Our $1$st approach \\ (with CNN II)}  & $93.25$ & $96.91$ & $90.62$ & $97.89$ & $86.74$ & $95.80$ & $81.75$ & $89.68$\\
	\hline
	\tabincell{l} {Our $1$st approach \\ (with CNN I \& II)}  & $93.97$ & $98.19$ & $92.87$ & $98.38$ & $87.73$ & $96.56$ & $84.13$ & $92.07$\\
	\hline
	\tabincell{l} {Our $2$st approach \\ (with global \\ features only)} & $90.50$ & $-$ & $91.15$ & $-$ & $83.98$ & $-$ & $-$ & $-$ \\
	\hline
	\tabincell{l} {Our $2$st approach \\ (with both local \\ and global features)} & $94.85$ & $-$ & $94.19$ & $-$ & $88.38$ & $-$ & $-$ & $-$ \\
	\hline
	\end{tabular}
	}

	\end{center}
\end{table*}

We also evaluated the recognition performance of our CNN I and CNN II models. CNN II do does not show obvious priority on these test data. However, the combination of CNN I and CNN II for character recognition gives much higher accuracy on both character level and plate level, which means that the features learned by CNN I and CNN II are complementary. The combined features can lead to better result.

\begin{figure}[tb]
\centering
\includegraphics[width=0.3\textwidth]{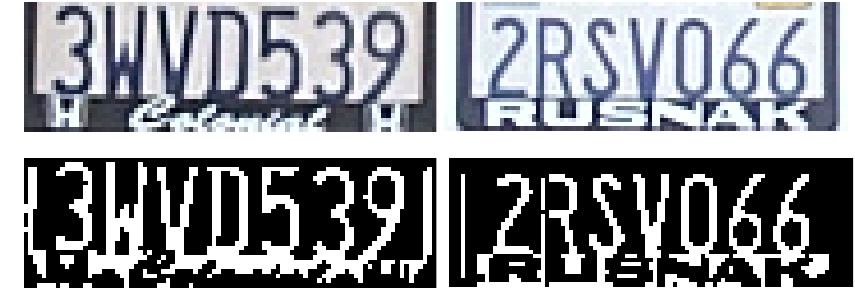}
\caption{License plates from Caltech cars dataset. The top line is the detected license plates, and the bottom line is the binarized results. The subtitles cause significant disturbance for CC based character segmentation, and lead to bad segmentation results.}
\label{Fig:6}
\end{figure}

It is noted that the recognition accuracy of the first method on Caltech cars dataset is not very high. That is mainly caused by the poor segmentation results. As the plate characters in Caltech cars license plate are  connected to the subtitle in Caltech cars license plates, CC based method cannot separate them well, which leads to the poor recognition results either. Some examples from the Caltech cars license plates are shown in Fig.~\ref{Fig:6}.

\begin{figure*}[t!]
\centering
\includegraphics[width=0.95\textwidth]{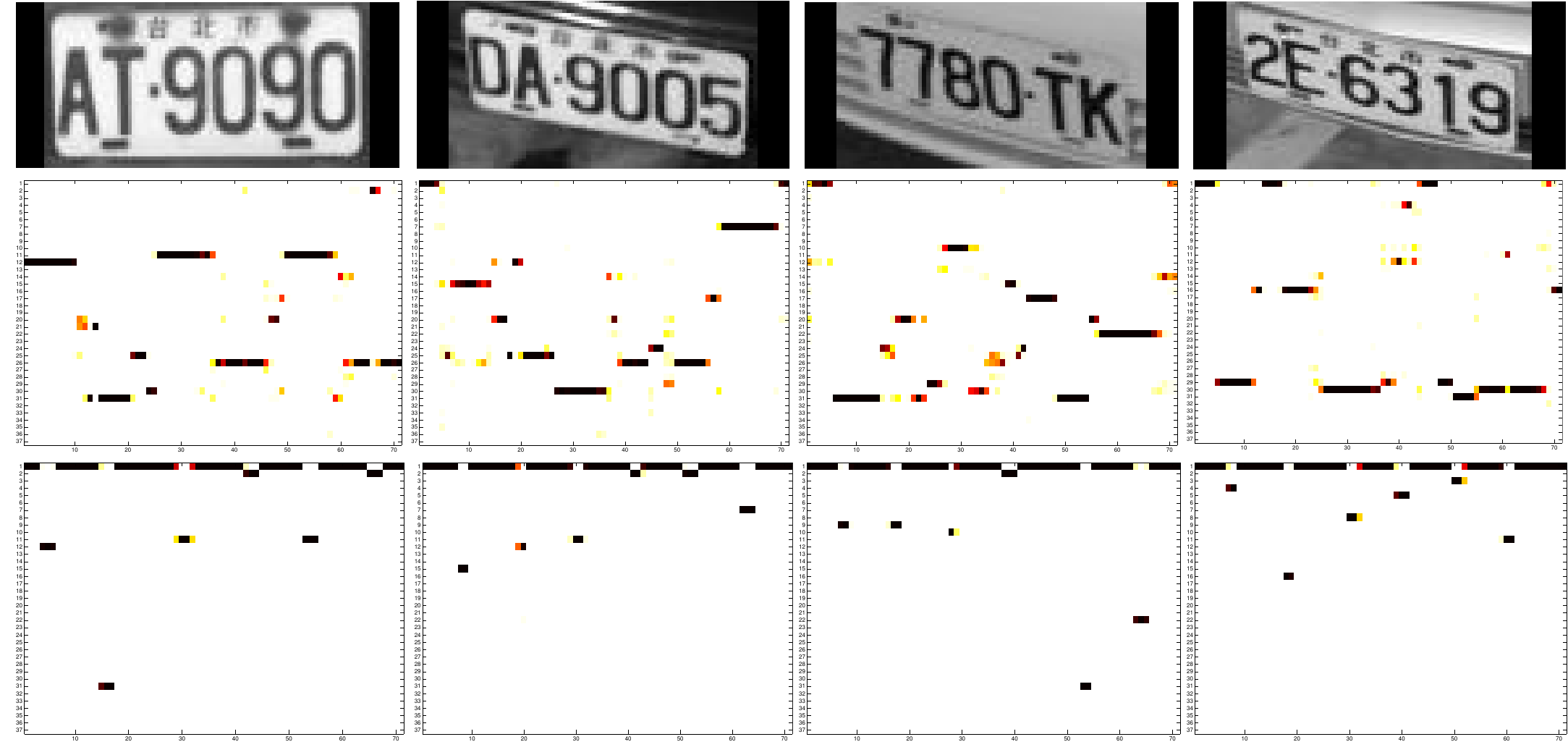}
\caption{License plate recognition confidence maps. The first row is the detected license plate. The second row is the recognition probabilities from the soft-max layer of CNN. The third row is the recognition probabilities from BRNN. For each confidence map, the recognition probabilities of current sub-window on $37$-class are shown vertically (with classes order from top to bottom: non-character, $0$-$9$, A-Z). BRNN gives better recognition results. Characters on each license plate can be read straightforward from the outputs of BRNN.}
\label{Fig:9}
\end{figure*}

Without character segmentation, our second approach achieves the highest recognition accuracy on AOLP dataset. The second method has not been applied to the Caltech cars dataset because we do not have training data with the similar pattern and distribution as Caltech cars license plates. For AOLP dataset, the experiments are carried out by using license plates from different sub-datasets for training and test separately. For example, we use the license plated from LE and RP sub-datasets to train the BRNN, and test its performance on AC sub-dataset. Similarly, AC and RP are used for training and LE for test, and so on.
Data augmentation is also implemented via image translation and affine transformation to reduce overfitting. Since the license plates in RP have a large degree of rotation and projective orientation, features extracted horizontally through sliding window are inaccurate for each character. Hence Hough transform is employed here to correct rotations~\cite{Rasheed2012}. Experimental results in the last row of Table~\ref{Tab:8} demonstrate the superiority of the sequence labelling based recognition method. It not only skips the challenging task of character separation, but also takes advantage of the abundant context information, which helps to enhance the recognition accuracy for each character. %
In order to show the advantage of BRNN further, we also visualize the recognition results from the soft-max layer of CNN and BRNN respectively. The $9$-layer CNN model is retrained by adding background images and using bootstrapping so that it can distinguish characters as well as background. The recognition probability distributions from the soft-max layer of CNN and BRNN are compared in Fig.~\ref{Fig:9}. %
It can be observed that the character recognition probabilities are more clear and correct on the output maps of BRNN. Characters can then be separated naturally, and the final license plate reading is straightforward by applying CTC on these maps.

\begin{figure*}[tb]
\centering
\includegraphics[width=0.95\textwidth]{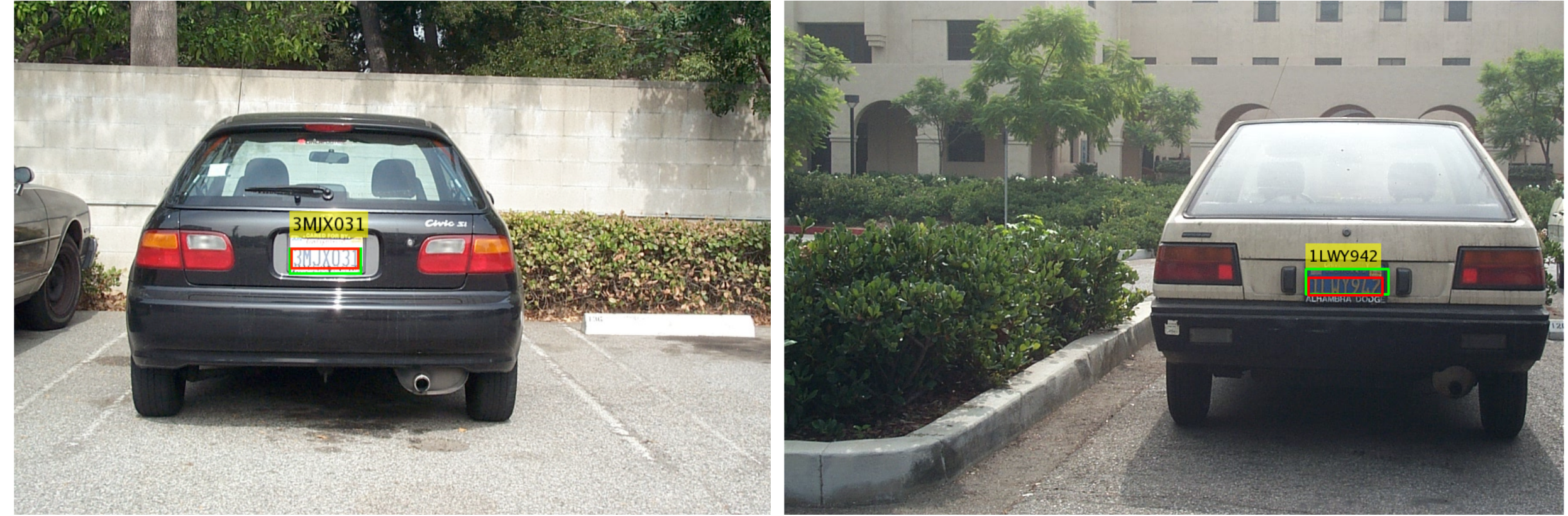}
\caption{Examples of license plate detection and recognition on the Caltech cars dataset. The red rectangles are the ground-truth,
while the green ones are our detection results. The yellow tags present the recognition results.}
\label{Fig:7}
\end{figure*}

\begin{figure*}[tb]
\centering
\includegraphics[width=0.75\textwidth]{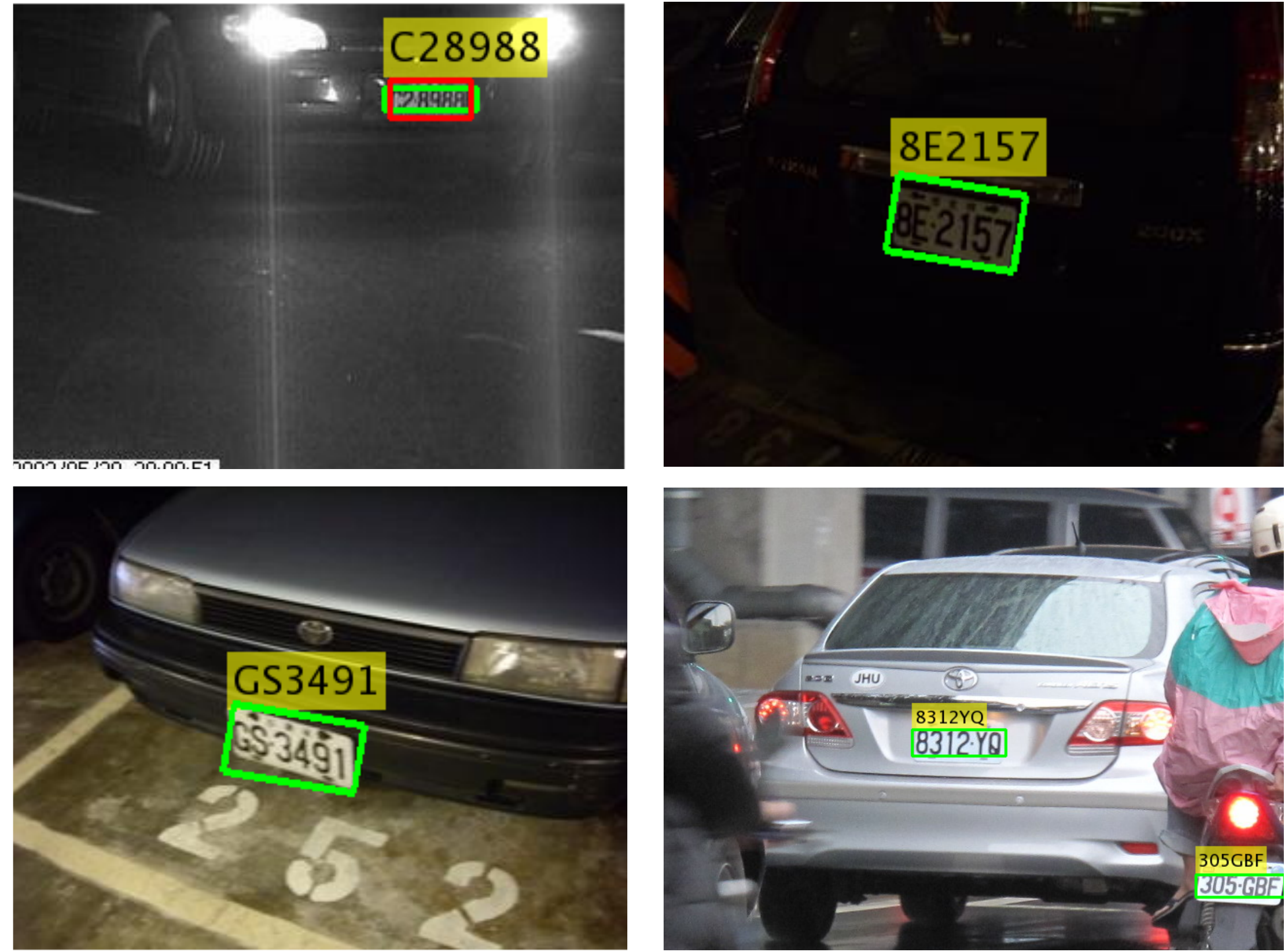}
\caption{Examples of license plate detection and recognition on the AOLP dataset. The green rectangles shows our detection results, with the recognition results presented in the yellow tags. The results demonstrate the robustness of our methods. It can detect and recognize license plates under various illuminations and orientations.}
\label{Fig:8}
\end{figure*}

Some examples of the license plates detection and recognition results are shown in Fig.~\ref{Fig:7} for Caltech cars dataset and in Fig.~\ref{Fig:8} for AOLP dataset.

\section{Conclusion}
 \label{SEC:Con} 

In this paper we have presented a license plate detection and recognition system using the promising CNN technique. We designed a simpler $4$-layer CNN and a deeper $9$-layer CNN for fast detection and accurate recognition respectively. The $4$-layer CNN is trained with $37$-class outputs, which learns specific features for each character, and is more effective to detect characters than a binary text/non-text classifier. The $9$-layer CNN with a much deeper architecture can learn more discriminate features which is robust to various illumination, rotations and distortions in the image, and lead to a higher recognition accuracy. Including LBP features into the input data can help to enhance the performance of CNN to some extent. The sequence labelling based method is able to recognize the whole license plate without character-level segmentation. The recurrent property of RNN enables it to explore context information, which contributes a lot to the final recognition result. Experimental result shows that this method can produce impressive performance given sufficient training data.

However, there are also some limitations in our work. The obvious one is on the efficiency. Although current detection speed is not unbearable with the aid of GPU, it still cannot be used in real time. Therefore, methods will be explored to improve the detection speed. One direction is to reduce the detection area using proposal based approaches.

\bibliographystyle{IEEEtran}
\bibliography{IEEEabrv,mybibfile}

\begin{thebibliography}{10}
\providecommand{\url}[1]{#1}
\csname url@samestyle\endcsname
\providecommand{\newblock}{\relax}
\providecommand{\bibinfo}[2]{#2}
\providecommand{\BIBentrySTDinterwordspacing}{\spaceskip=0pt\relax}
\providecommand{\BIBentryALTinterwordstretchfactor}{4}
\providecommand{\BIBentryALTinterwordspacing}{\spaceskip=\fontdimen2\font plus
\BIBentryALTinterwordstretchfactor\fontdimen3\font minus
  \fontdimen4\font\relax}
\providecommand{\BIBforeignlanguage}[2]{{%
\expandafter\ifx\csname l@#1\endcsname\relax
\typeout{** WARNING: IEEEtran.bst: No hyphenation pattern has been}%
\typeout{** loaded for the language `#1'. Using the pattern for}%
\typeout{** the default language instead.}%
\else
\language=\csname l@#1\endcsname
\fi
#2}}
\providecommand{\BIBdecl}{\relax}
\BIBdecl

\bibitem{Du2013Automatic}
S.~Du, M.~Ibrahim, M.~Shehata, and W.~Badawy, ``Automatic license plate
  recognition (alpr): A state-of-the-art review,'' \emph{{IEEE} Trans. Circuits
  Syst. Video Technol.}, vol.~23, no.~2, pp. 311--325, 2013.

\bibitem{Cnagnost2008}
C.~Anagnostopoulos, I.~Anagnostopoulos, I.~Psoroulas, V.~Loumos, and
  E.~Kayafas, ``License plate recognition from still images and video
  sequences: A survey,'' \emph{{IEEE} Trans. Intell. Transp. Syst.}, vol.~9,
  no.~3, pp. 377--391, 2008.

\bibitem{Ross2014}
\BIBentryALTinterwordspacing
R.~Girshick, J.~Donahue, T.~Darrell, and J.~Malik, ``Rich feature hierarchies
  for accurate object detection and semantic segmentation,'' in \emph{Proc.
  IEEE Conf. Comp. Vis. Patt. Recogn.}, 2014. [Online]. Available:
  \url{http://arxiv.org/abs/1311.2524}
\BIBentrySTDinterwordspacing

\bibitem{Graves2009Pami}
A.~Graves, M.~Liwicki, and S.~Fernandez, ``A novel connectionist system for
  unconstrained handwriting recognition,'' \emph{{IEEE} Trans. Pattern Anal.
  Mach. Intell.}, vol.~31, no.~5, pp. 855--868, 2009.

\bibitem{Zhou2012Principal}
W.~Zhou, H.~Li, Y.~Lu, and Q.~Tian, ``Principal visual word discovery for
  automatic license plate detection,'' \emph{{IEEE} Trans. Image Process.},
  vol.~21, no.~9, pp. 4269--4279, 2012.

\bibitem{Anagnostopoulos}
C.~Anagnostopoulos, I.~Anagnostopoulos, V.~Loumos, and E.~Kayafas, ``A license
  plate-recognition algorithm for intelligent transportation system
  applications,'' \emph{{IEEE} Trans. Intell. Transp. Syst.}, vol.~7, no.~3,
  pp. 377--392, 2006.

\bibitem{Bai2004}
H.~Bai and C.~Liu, ``A hybrid license plate extraction method based on edge
  statistics and morphology,'' in \emph{Proc. IEEE Int. Conf. Patt. Recogn.},
  2004, pp. 831--834.

\bibitem{Qiu2009}
Y.~Qiu, M.~Sun, and W.~Zhou, ``License plate extraction based on vertical edge
  detection and mathematical morphology,'' in \emph{Proc. Int. Conf. Comp.
  Intell. Softw. Engin.}, 2009, pp. 1--5.

\bibitem{Zheng2005}
D.~Zheng, Y.~Zhao, and J.~Wang, ``An efficient method of license plate
  location,'' \emph{Pattern Recogn. Lett.}, vol.~26, no.~15, pp. 2431--2438,
  2005.

\bibitem{Tan2013}
J.~Tan, S.~Abu-Bakar, and M.~Mokji, ``License plate localization based on
  edge-geometrical features using morphological approach,'' in \emph{Proc. IEEE
  Int. Conf. Image Process.}, 2013, pp. 4549--4553.

\bibitem{Lalimi2013}
M.~A. Lalimia, S.~Ghofrania, and D.~McLernonb, ``A vehicle license plate
  detection method using region and edge based methods,'' \emph{Comp \& Electr.
  Engin.}, vol.~39, p. 834–845, 2013.

\bibitem{Chen2012}
R.~Chen and Y.~Luo, ``An improved license plate location method based on edge
  detection,'' in \emph{Proc. Int. Conf. Appl. Phys. Industr. Engin.}, 2012, p.
  1350–1356.

\bibitem{Rasheed2012}
S.~Rasheed, A.~Naeem, and O.~Ishaq, ``Automated number plate recognition using
  hough lines and template matching,'' in \emph{Proc. World Cong. Engin. Comp.
  Sci.}, 2012, pp. 199--203.

\bibitem{Deb2008}
K.~Deb and K.~Jo, ``Hsi color based vehicle license plate detection,'' in
  \emph{Proc. Int. Conf. Cont. Autom. Syst.}, 2008, pp. 687--691.

\bibitem{Jia2007}
W.~Jia, H.~Zhang, and X.~He, ``Region-based license plate detection,''
  \emph{Jour. Netw. Comp. Appl.}, vol.~30, pp. 1324--1333, 2007.

\bibitem{Zhang2006}
H.Zhang, W.Jia, X.He, and Q.Wu, ``Learning-based license plate detection using
  global and local features,'' in \emph{Proc. IEEE Int. Conf. Patt. Recogn.},
  2006, pp. 1102--1105.

\bibitem{Giannoukos2006}
C.-N. Anagnostopoulos, I.~Giannoukos, V.~Loumos, and E.~Kayafas, ``A license
  plate recognition algorithm for intelligent transportation system
  applications,'' \emph{{IEEE} Trans. Intell. Transp. Syst.}, vol.~7, no.~3,
  pp. 377--392, 2006.

\bibitem{Giannoukos2010}
I.~Giannoukos, C.-N. Anagnostopoulos, V.~Loumos, and E.~Kayafas, ``Operator
  context scanning to support high segmentation rates for real time license
  plate recognition,'' \emph{Pattern Recogn.}, vol.~43, no.~11, p. 3866–3878,
  2010.

\bibitem{Yu2015}
S.~Yu, B.~Li, Q.~Zhang, C.~Liu, and M.~Meng, ``A novel license plate location
  method based on wavelet transform and emd analysis,'' \emph{Pattern Recogn.},
  vol.~48, no.~1, p. 114–125, 2015.

\bibitem{Lin2010}
K.~Lin, H.~Tang, and T.~Huang, ``Robust license plate detection using image
  saliency,'' in \emph{Proc. IEEE Int. Conf. Patt. Recogn.}, 2010, pp.
  3945--3948.

\bibitem{Li2013}
B.~Li, B.~Tian, Y.~Li, and D.~Wen, ``Component-based license plate detection
  using conditional random field model,'' \emph{{IEEE} Trans. Intell. Transp.
  Syst.}, vol.~14, no.~4, pp. 1690--1699, 2013.

\bibitem{Nomura2005}
S.~Nomura, K.~Yamanaka, O.~Katai, H.~Kawakami, and T.~Shiose, ``A novel
  adaptive morphological approach for degraded character image segmentation,''
  \emph{Pattern Recogn.}, vol.~38, pp. 1961--1975, 2005.

\bibitem{Guo2008}
J.~Guo and Y.~Liu, ``License plate localization and character segmentation with
  feedback self-learning and hybrid binarization techniques,'' \emph{{IEEE}
  Trans. Veh. Technol.}, vol.~57, no.~3, pp. 1417--1424, 2008.

\bibitem{Qiao2010}
S.~Qiao, Y.~Zhu, X.~Li, T.~Liu, and B.~Zhang, ``Research on improving the
  accuracy of license plate character segmentation,'' in \emph{Proc. Int. Conf.
  Front. Comp. Sci. Tech.}, 2010, pp. 489--493.

\bibitem{Chang2004}
S.~Chang, L.~Chen, Y.~Chung, and S.~Chen, ``Automatic license plate
  recognition,'' \emph{{IEEE} Trans. Intell. Transp. Syst.}, vol.~5, no.~1, p.
  42–53, 2004.

\bibitem{Jiao2009}
J.~Jiao, Q.~Ye, and Q.~Huang, ``A configurable method for multi-style license
  plate recognition,'' \emph{Pattern Recogn.}, vol.~42, pp. 358--369, 2009.

\bibitem{Zheng2013An}
L.~Zheng, X.~He, B.~Samali, and L.~Yang, ``An algorithm for accuracy
  enhancement of license plate recognition,'' \emph{J. Comp. \& Syst. Sci.},
  vol.~79, no.~2, pp. 245--255, 2013.

\bibitem{Zhang2013}
Y.~Zhang, Z.~Zha, and L.~Bai, ``A license plate character segmentation method
  based on character contour and template matching,'' \emph{Applied Mechanics
  and Materials}, vol. 333-335, pp. 974--979, 2013.

\bibitem{Capar2006}
A.~Capar and M.~Gokmen, ``Concurrent segmentation and recognition with
  shape-driven fast marching methods,'' in \emph{Proc. IEEE Int. Conf. Patt.
  Recogn.}, 2006, pp. 155--158.

\bibitem{Goel2013}
S.~Goel and S.~Dabas, ``Vehicle registration plate recognition system using
  template matching,'' in \emph{Proc. Int. Conf. Signal Proc. Communication},
  2013, pp. 315--318.

\bibitem{Ko2003}
M.~Ko and Y.~Kim, ``License plate surveillance system using weighted template
  matching,'' in \emph{Proc. $32$nd Applied Imagery Patt. Recog. Workshop},
  2003, pp. 269--274.

\bibitem{Llorens2005}
D.~Llorens, A.~Marzal, V.~Palazon, and J.~M. Vilar, ``Car license plates
  extraction and recognition based on connected components analysis and hmm
  decoding,'' \emph{Lecture Notes in Computer Science}, vol. 3522, pp.
  571--578, 2005.

\bibitem{Wen2011}
Y.~Wen, Y.~Lu, J.~Yan, Z.~Zhou, K.~von Deneen, and P.~Shi, ``An algorithm for
  license plate recognition applied to intelligent transportation system,''
  \emph{{IEEE} Trans. Intell. Transp. Syst.}, vol.~12, pp. 830--845, 2011.

\bibitem{Liu2010}
L.~Liu, H.~Zhang, A.~Feng, X.~Wang, and J.~Guo, ``Simplified local binary
  pattern descriptor for character recognition of vehicle license plate,'' in
  \emph{Proc. Int. Conf. Comp. Graph. Imag. Visual.}, 2010, pp. 157--161.

\bibitem{Sharma2014}
J.~Sharma, A.~Mishra, K.~Saxena, and S.~Kumar, ``A hybrid technique for license
  plate recognition based on feature selection of wavelet transform and
  artificial neural network,'' in \emph{Proc. Int. Conf. Optim. Reliab. Infor.
  Techn.}, 2014, pp. 347--352.

\bibitem{Wang2012}
T.~Wang, D.~Wu, A.~Coates, and A.~Y. Ng, ``End-to-end text recognition with
  convolutional neural networks,'' in \emph{Proc. IEEE Int. Conf. Patt.
  Recogn.}, 2012, pp. 3304--3308.

\bibitem{Max2014ECCV}
M.~Jaderberg, A.~Vedaldi, and A.~Zisserman, ``Deep features for text
  spotting,'' in \emph{Proc. Eur. Conf. Comp. Vis.}, 2014, pp. 512--528.

\bibitem{Graves2006ICML}
A.~Graves, S.~Fernandez, F.~Gomez, and J.~Schmidhuber, ``Connectionist temporal
  classification: Labelling unsegmented sequence data with recurrent neural
  networks,'' in \emph{Proc. Int. Conf. Mach. Learn.}, 2006, pp. 369--376.

\bibitem{Sutskever2014}
\BIBentryALTinterwordspacing
I.~Sutskever, O.~Vinyals, and Q.~V. Le, ``Sequence to sequence learning with
  neural networks,'' in \emph{Proc. Neural Infor. Proc. Syst.}, 2014. [Online].
  Available: \url{http://arxiv.org/abs/1409.3215}
\BIBentrySTDinterwordspacing

\bibitem{Su2014ACCV}
B.~Su and S.~Lu, ``Accurate scene text recognition based on recurrent neural
  network,'' in \emph{Proc. Asi. Conf. Comp. Vis.}, 2014, pp. 35--48.

\bibitem{He2015Reading}
\BIBentryALTinterwordspacing
P.~He, W.~Huang, Y.~Qiao, C.~C. Loy, and X.~Tang, ``Reading scene text in deep
  convolutional sequences,'' \emph{Technical report}, 2015. [Online].
  Available: \url{http://arxiv.org/abs/1506.04395}
\BIBentrySTDinterwordspacing

\bibitem{Milyaev2015}
S.~Milyaev, O.~Barinova, T.~Novikova, P.~Kohli, and V.~Lempitsky, ``Fast and
  accurate scene text understanding with image binarization and off-the-shelf
  ocr,'' \emph{Int. Jour. Doc. Anal. Recog.}, vol.~18, pp. 169--182, 2015.

\bibitem{Yoon2011}
Y.~Yoon, K.~Ban, H.~Yoon, and J.~Kim, ``Blob extraction based character
  segmentation method for automatic license plate recognition system,'' in
  \emph{Proc. IEEE Int. Conf. System, Man, Cybernetics}, 2011, pp. 2192--2196.

\bibitem{Zhong2015High}
\BIBentryALTinterwordspacing
Z.~Zhong and Z.~X. L.~Jin, ``High performance offline handwritten chinese
  character recognition using {GoogLeNet} and directional feature maps,'' in
  \emph{Proc. Int. Conf. Doc. Anal. Recog.}, 2015. [Online]. Available:
  \url{http://arxiv.org/abs/1505.04925}
\BIBentrySTDinterwordspacing

\bibitem{Hsu2013}
G.~Hsu, J.~Chen, and Y.~Chung, ``Application-oriented license plate
  recognition,'' \emph{{IEEE} Trans. Veh. Technol.}, vol.~62, no.~2, pp.
  552--561, 2013.

\bibitem{Chen2011LBP}
X.~Chen and C.~Qi, ``A super-resolution method for recognition of license plate
  character using {LBP} and {RBF},'' in \emph{Proc. Int. Work. Mach. Learn. for
  Signal Process.}, 2011, pp. 1--5.

\bibitem{vedaldi15matconvnet}
A.~Vedaldi and K.~Lenc, ``{MatConvNet}---convolutional neural networks for
  {MATLAB},'' in \emph{Proc. {ACM} Int. Conf. Multimedia}, 2015.

\bibitem{Wangkai2011}
K.~Wang, B.~Babenko, and S.~Belongie, ``End-to-end scene text recognition,'' in
  \emph{Proc. IEEE Int. Conf. Comp. Vis.}, 2011.

\bibitem{icdar2003}
S.~M. Lucas, A.~Panaretos, L.~Sosa, A.~Tang, S.~Wong, and R.~Young, ``{ICDAR}
  2003 robust reading competitions,'' in \emph{Proc. Int. Conf. Doc. Anal.
  Recog.}, 2003, pp. 682--687.

\bibitem{icdar2005}
S.~Lucas, ``{ICDAR} 2005 text locating competition results,'' in \emph{Proc.
  Int. Conf. Doc. Anal. Recog.}, 2005, pp. 80--84.

\bibitem{icdar2011}
A.~Shahab, F.~Shafait, and A.~Dengel, ``{ICDAR} 2011 robust reading competition
  challenge 2: Reading text in scene images,'' in \emph{Proc. Int. Conf. Doc.
  Anal. Recog.}, 2011, pp. 1491--1496.

\bibitem{msrcorid}
\BIBentryALTinterwordspacing
A.~Criminisi. (2004) Microsoft research cambridge object recognition image
  database. [Online]. Available:
  \url{http://research.microsoft.com/en-us/downloads/b94de342-60dc-45d0-830b-9f6eff91b301/default.aspx}
\BIBentrySTDinterwordspacing

\bibitem{carmarkus}
\BIBentryALTinterwordspacing
(2003) Caltech plate dataset. [Online]. Available:
  \url{http://www.vision.caltech.edu/html-files/archive.html}
\BIBentrySTDinterwordspacing

\bibitem{Le2006}
W.~Le and S.~Li, ``A hybrid license plate extraction method for complex
  scenes,'' in \emph{Proc. IEEE Int. Conf. Patt. Recogn.}, 2006, pp. 324--327.

\bibitem{Wooi2010}
H.~W. Lim and Y.~H. Tay, ``Detection of license plate characters in natural
  scene with {MSER} and {SIFT} unigram classifier,'' in \emph{Proc. IEEE Int.
  Conf. Sustainable Utilization and Development in Engineering and Technology},
  2010, pp. 95--98.

\end{thebibliography}

\end{document}